\documentclass{article}
\usepackage[utf8]{inputenc}
\usepackage{amsmath,graphicx,stmaryrd,subfig,color,authblk, url, makecell}

\title{A Study for Universal Adversarial Attacks on Texture Recognition}

\author[1]{Yingpeng Deng}
\author[1,2]{Lina J. Karam}
\affil[1]{Image, Video and Usability Lab, School of ECEE, Arizona State University, Tempe, AZ, USA}
\affil[2]{Dept. of ECE, School of Engineering, Lebanese American University, Lebanon}
\affil[ ]{\{ypdeng, karam\}@asu.edu}
\date{}

\begin{document}

\maketitle

\begin{abstract}
	Given the outstanding progress that convolutional neural networks (CNNs) have made on natural image classification and object recognition problems, it is shown that deep learning methods can achieve very good recognition performance on many texture datasets. However, while CNNs for natural image classification/object recognition tasks have been revealed to be highly vulnerable to various types of adversarial attack methods, the robustness of deep learning methods for texture recognition is yet to be examined. In our paper, we show that there exist small image-agnostic/univesal perturbations that can fool the deep learning models with more than 80\% of testing fooling rates on all tested texture datasets. The computed perturbations using various attack methods on the tested datasets are generally quasi-imperceptible, containing structured patterns with low, middle and high frequency components.
\end{abstract}

\section{Introduction}

Texture is one of the essential features for objects or identities in the physical world. It usually yields visually distinguishable properties to provide mid-level hints for human/machine vision. For example, Convolutional neural networks (CNNs)~\cite{Simonyan15,szegedy2015going,szegedy2016rethinking,he2016deep} are regarded as highly effective solutions to ImageNet ILSVRC dataset~\cite{ILSVRC15}, a challenging large-scale image classification dataset containing more than one million of real-world images categorized into one thousand classes. In the inference process of CNNs, local features are collected and combined by convolutional filter banks from previous layers, where textures in the input image are the valuable local information to collect for the convolutional filters in the first few layers. Moreover, a recent study has shown that ImageNet-trained CNNs rely on textures rather than shapes to recognize images/objects~\cite{geirhos2018imagenet}, which can indicate that textures play a pivotal role in the success of the CNN models for object recognition.

In fact, texture recognition is already a widely explored topic. Before the prevalence of CNNs, researchers try to extract or design discriminative features from texture patterns via various kinds of methods, such as Scale-Invariant Feature Transform (SIFT)~\cite{lowe2004distinctive}, Bag-of-Words (BoWs)~\cite{csurka2004visual}, Vector of Locally Aggregated Descriptors (VLAD)~\cite{jegou2010aggregating}, Fisher Vector (FV)~\cite{perronnin2010improving} and so forth. Given the superior performance of CNNs on object recognition problems since the early 2010s~\cite{krizhevsky2012imagenet}, Cimpoi \textit{et al.}~\cite{cimpoi2015deep} took advantage of pretrained CNN features and processed them with different encoders to achieve cutting-edge performance on texture recognition problems. Furthermore, inspired by dictionary learning, Zhang \textit{et al.}~\cite{zhang2017deep} proposed a Deep Texture Encoding Network (DeepTEN), making end-to-end learning possible, followed by some variants to improve DeepTEN~\cite{xue2018deep, hu2019multi}.

Despite the outstanding performances of CNNs, some researchers have casted doubt on their robustness against small perturbations~\cite{szegedy2013intriguing}. Specifically, it has been shown that there exist small and image-agnostic perturbations, called universal perturbations, that can fool CNN classifiers, resulting in a significant decrease in classification accuracy over a natural image set~\cite{moosavi2017universal}. However, to the best of our knowledge, the performance of deep neural networks (DNNs) under universal adversarial attacks has not been studied for texture recognition. Given the properties of repetitive/redundant information and spatially invariant representation in texture images, what is the effect of universal perturbations on the performance of DNNs for texture image recognition? Can the performance be severely degraded by universal perturbations that are computed with much less training images and classes than those computed on ImageNet? Do the adopted modules and architectures in the deep learning models for texture recognition really improve the ability and robustness of the baseline DNN models? In order to answer these questions, we train four available deep learning models on six texture datasets and conduct adversarial attack experiments with existing universal attack algorithms. We perform the universal attacks with the assumption that we have full knowledge about the targeted model (white-box attack), such as its architecture, weight and gradient information, and that relevant training data is available (data-dependent). After the computation of the perturbation, we use testing data to evaluate the attack performance. The rest of our paper is organized as follows. Section 2 presents an overview of related work on texture recognition and adversarial attacks. Our experimental setup and obtained performance results are described in Section 3. Finally a conclusion is given in Section 4.

\section{Related Work}

\subsection{Texture Recognition}

Here we only consider the end-to-end deep-learning models for texture recognition, which most of the existing universal attack methods were proposed to fool. Inspired by previous dictionary learning and residual encoding techniques such as VLAD and Fisher Vector, Zhang \textit{et al.}~\cite{zhang2017deep} proposed a learnable residual encoding layer, where dictionary codewords and assigned weights can both be updated via backpropagation, and plugged it before the fully connected layer of the ResNet structure. Further, to take the local spatial information into consideration, Xue \textit{et al.}~\cite{xue2018deep} combined the features after the global average pooling layer with the outputs of the encoding layer proposed in~\cite{zhang2017deep} through a bilinear model~\cite{freeman1997learning} in their deep encoding pooling (DEP) network. Recently, Hu \textit{et al.}~\cite{hu2019multi} presented the idea of multi-level texture encoding and representation (MuLTER) by assembling multi-stage features extracted using the DEP module after each residual block before the final fully connected layer. Intuitively, all these architectures are derived from the vanilla ResNet architecture.

\subsection{Universal Attacks}

White-box universal perturbations that are computed using relevant training data can produce the stronger attack ability than other kinds of universal perturbations such as data-independent~\cite{liu2019universal} and black-box ones~\cite{tsuzuku2019structural}. Based on the image-dependent attack algorithm DeepFool~\cite{moosavi2016deepfool}, Moosavi-Dezfooli \textit{et al}~\cite{moosavi2017universal} produced universal adversarial perturbations (UAP) by accumulating the image-dependent updates iteratively over the training images to fool the targeted CNN model. Generative adversarial networks were adopted to generate the universal perturbation in~\cite{poursaeed2018generative, reddy2018nag}, where the adversarial networks were set as the targeted models. Poursaeed \textit{et al}~\cite{poursaeed2018generative} trained the generative network using a fixed random pattern as input to produce the generative adversarial perturbation (GAP), while Mopuri \textit{et al}~\cite{reddy2018nag} introduced both fooling and diversity objectives as the loss function of their network for adversary generation (NAG) to learn the perturbation distribution in the latent space. Recently, the authors of~\cite{mummadi2019defending} and~\cite{shafahi2020universal} used mini-batch based stochastic projected gradient descent (sPGD) during training by maximizing the average loss over each mini-batch, enlighted by another image-dependent attack algorithm called iterative gradient sign method (IGSM) or projected gradient descent (PGD)~\cite{kurakin2016adversarial}. Most recently, Deng and Karam ~\cite{9191288} proposed an enhanced universal projected gradient descent (UPGD) based on the UAP framework by replacing DeepFool with the stronger PGD attack and boosting the computation process with momentum.

\section{Experiments}

We consider six texture datasets in order to examine the recognition performance of the deep learning models under various unviersal attacks. The Materials in Context (MINC) Database~\cite{bell2015material} is a large real-world material dataset. In our work, we adopt its publicly available subset MINC-2500 with provided train-test split. There are 23 classes with 2500 images for each class. Xue \textit{et al.} created the Ground Terrain in Outdoor Scenes (GTOS)~\cite{xue2017differential} dataset with 31 classes of over 90,000 ground terrain images and the GTOS-mobile ~\cite{xue2018deep} with the same classes but with a much smaller number of images (around 6,000 images). As in~\cite{xue2018deep}, we adopt the GTOS dataset as the training set and test the trained models on the GTOS-mobile dataset. In the rest of our paper, we use MINC and GTOS to refer to the MINC-2500 and the combined dataset of GTOS and GTOS-mobile, respectively.

We also adopt some smaller texture datasets. The Describable Textures Database (DTD)~\cite{cimpoi2014describing} includes 47 categories with 120 images per category. The 4D light-field (4DLF) material dataset~\cite{wang20164d} consists of 1200 images in total for 12 different categories. An angular resolution of $7 \times 7$ is used for each image in the dataset and we only use the one where $(u, v) = (-3, 3)$ in our experiments. The Filckr Material Dataset (FMD)~\cite{sharan2013recognizing} has 10 material classes and 100 images per class. The KTH-TIPS-2b (KTH)~\cite{caputo2005class} comprises 11 texture classes, with four samples per class and 108 images per sample.

We evaluate the performance of ResNet~\cite{he2016deep}, DeepTEN~\cite{zhang2017deep}, DEP~\cite{xue2018deep} and MuLTER~\cite{hu2019multi} on the six texture datasets. For each dataset, the training set is used to train the models for texture recognition, the attacking set includes images that are randomly sampled from the training set for computing the perturbations, and the testing set is to evalute the performance without and with universal attacks. The number of images for each purpose is listed in Table~\ref{imgnums}. We use the train-test split that is either provided in the dataset or suggested in~\cite{zhang2017deep}. To extract the attacking set, we randomly sample 10\% of the images in the training image set for the GTOS/KTH dataset, one-third of the training images for the FMD dataset and a number of training images that is equal to the number of testing images for MINC, DTD and 4DLF separately.

\begin{table*}
	\centering
	\caption{The numbers of data used for training the networks (training set), computing the perturbation (attacking set) and testing (testing set).}
	\label{imgnums}
	\begin{tabular}[t]{|c|c|c|c|}
		\hline
		Dataset&training set&attacking set&testing set\\
		\hline
		MINC~\cite{bell2015material}&48,875&5,750&5,750\\
		\hline
		GTOS~\cite{xue2017differential, xue2018deep}&93,945&9,381&6,066\\
		\hline
		DTD~\cite{cimpoi2014describing}&3,760&1,880&1,880\\
		\hline
		4DLF~\cite{wang20164d}&840&360&360\\
		\hline
		FMD~\cite{sharan2013recognizing}&900&300&100\\
		\hline
		KTH~\cite{caputo2005class}&2,376&231&2,376\\
		\hline
	\end{tabular}
\end{table*}

\subsection{Implementations}

We train the four deep learning models on each of the six texture datasets and then conduct the mentioned universal attacks on 24 trained classifiers. Similar to the training process of deep learning models, the perturbations are computed only on each of the attacking sets, whose images are randomly sampled from the corresponding training set, and then are evaluated on the testing data.

\textbf{Training.} The deep learning models are finetuned via transfer learning based on the ResNet backbones which are pretrained on ImageNet. We follow the suggestions of backbones on different datasets in~\cite{zhang2017deep, xue2018deep}. In regard to the network backbone, we use pretrained ResNet50\footnote{According to~\cite{pytorch-encoding}, the adopted ResNet50 backbones are slightly different from~\cite{he2016deep}, where the first convolutional layer with a kernel size of 7 is replaced by three cascaded 3 $\times$ 3 convolutional layers.} for MINC, DTD, 4DLF and FMD and pretrained ResNet18 for GTOS and KTH. Following the same training strategies in~\cite{xue2018deep, hu2019multi}, we train our models with only single-size images. Similar to the data augmentation strategies in~\cite{zhang2017deep}, the input images are resized to 256 $\times$ 256 and then randomly cropped to 224 $\times$ 224, followed by a random horizontal flipping. Standard color augmentation and PCA-based noise are used as in~\cite{krizhevsky2012imagenet}. For DeepTEN, DEP and MuLTER, the number of codewords is set to 8 for the ResNet18 backbone and to 32 for the ResNet50 backbone. The learning rate is initialized to 0.01 and decays every 10 epochs by a factor of 0.1, with a momentum coefficient of 0.9. The training process stops after 30 epochs.

\textbf{Attacking.} Generally, no data augmentation is used while computing the universal perturbations. But in our implementation, given the small number of data for 4DLF, FMD and KTH, we use random cropping and horizontal flipping to prevent overfitting problems for perturbations. We perform UAP~\cite{moosavi2017universal}, GAP~\cite{poursaeed2018generative}, sPGD~\cite{shafahi2020universal} and UPGD~\cite{9191288} on the trained models. In~\cite{poursaeed2018generative}, the authors mention two ways to optimize the perturbation - (1) minimizing the least-likely class loss between the original prediction with the least probability and the perturbed prediction; (2) maximizing the orginal loss between the perturbation prediction and the true target. We refer to the former as GAP-llc and to the latter as GAP-tar. For GAP and sPGD, we use a mini-batch size of 32 when computing the perturbation. The defaulted parameters are used for all the considered attack methods, except for UPGD whose hyperparameters (i.e., initial learning rate and decay factor, momentum, etc.) were varied to maximize the performance for the computed perturbations in terms of fooling rate. Typically, we stop the training process of the perturbation and report the results when the fooling rates are unable to increase by more than 0.5\% within 5 epochs for all the attack methods. For UAP, we choose its best results in terms of fooling rate within 20 epochs given that its learning curves were observed to fluctuate drastically. All perturbations are constrained within an $l_\infty$ norm of 10 on 8-bit images (about 0.04 for the normalized image scale $[0, 1]$).

\subsection{Results}

\subsubsection{Performance Comparison Results Without Attacks}

We show the classification accuracies of each model without attacks on the six texture datasets in Table~\ref{clsacc}. For MINC and GTOS datasets, it seems that the modifications based on the ResNet backbone that were proposed in DeepTEN, DEP and MuLTER result in trivial improvements or even worse performances as compared to the baseline ResNet model. The residual encoding layer in DeepTEN significantly improves the classification performances on 4DLF, FMD and KTH datasets but fails on other larger datasets. Similarly, as compared to DeepTEN, DEP only shows higher classification accuracies on DTD and 4DLF. The core idea of MuLTER is the ensemble of multi-level features, which is built upon the DEP architecture. It presents an overall increase of top-1 accuracies based on the baseline ResNet model except for the DTD dataset.

\begin{table*}
	\centering
	\caption{Top-1 accuracies on the testing sets for texture recognition. Bold number in each column indicates best performance on the corresponding dataset.}
	\label{clsacc}
	\begin{tabular}[t]{|c|c|c|c|c|c|c|}
		\hline
		&MINC&GTOS&DTD&4DLF&FMD&KTH\\
		\hline
		ResNet&81.5&\textbf{80.1}&71.4&72.5&70.0&79.9\\
		DeepTEN&81.7&77.8&61.2&79.2&82.0&\textbf{83.5}\\
		DEP&81.5&77.5&\textbf{73.0}&82.2&81.0&79.9\\
		MuLTER&\textbf{81.8}&77.9&71.6&\textbf{83.6}&\textbf{84.0}&80.6\\
		\hline
	\end{tabular}
\end{table*}

\subsubsection{Attacking Results} 
\label{attack}

\begin{table*}
	\centering
	\caption{Fooling rates by the universal attacks against different models. Bold number indicates best performance on the corresponding model in each column within the same dataset and italic number denotes best performance on the corresponding dataset over different models and attack methods.}
	\label{fr}
	
	\resizebox{\textwidth}{!}{
	\begin{tabular}[t]{|c|cccc|c|cccc|c|}
		\hline
		&ResNet&DeepTEN&DEP&MuLTER&mean&ResNet&DeepTEN&DEP&MuLTER&mean\\
		\cline{2-11}
		&\multicolumn{5}{c|}{MINC}&\multicolumn{5}{c|}{GTOS}\\
		\hline
		UAP&86.6&73.1&59.8&66.4&71.5&51.0&27.7&31.6&39.0&37.3\\
		GAP-llc&74.9&84.4&87.1&80.3&81.7&45.9&49.6&57.6&54.6&51.9\\
		GAP-tar&\textbf{94.1}&\textbf{\underline{94.8}}&\textbf{94.5}&\textbf{94.2}&\textbf{94.4}&69.5&\textbf{\underline{81.5}}&\textbf{72.5}&\textbf{79.0}&75.6\\
		sPGD&93.8&94.3&93.1&93.4&93.7&51.8&74.9&70.0&76.3&68.3\\
		UPGD&93.4&93.7&93.1&93.7&93.5&\textbf{78.0}&77.5&72.4&77.8&\textbf{76.4}\\
		\hline
		&\multicolumn{5}{c|}{DTD}&\multicolumn{5}{c|}{4DLF}\\
		\hline
		UAP&26.2&36.6&32.4&38.9&33.5&83.6&72.5&79.2&77.2&78.1\\
		GAP-llc&55.7&75.3&61.4&76.0&67.1&69.7&75.6&81.1&72.2&74.7\\
		GAP-tar&\textbf{72.5}&\textbf{\underline{86.2}}&\textbf{81.1}&83.6&\textbf{80.9}&87.5&84.2&\textbf{\underline{90.0}}&\textbf{89.4}&\textbf{87.8}\\
		sPGD&70.9&83.9&78.8&\textbf{84.5}&79.5&86.4&84.4&85.6&81.7&84.5\\
		UPGD&71.8&82.6&80.5&79.9&78.7&\textbf{88.3}&\textbf{88.3}&86.1&81.9&86.2\\
		\hline
		&\multicolumn{5}{c|}{FMD}&\multicolumn{5}{c|}{KTH}\\
		\hline
		UAP&65.0&49.0&39.0&50.0&50.8&75.8&41.9&62.0&42.0&55.4\\
		GAP-llc&51.0&67.0&78.0&55.0&62.8&48.2&57.6&8.4&45.1&39.8\\
		GAP-tar&\textbf{90.0}&\textbf{\underline{94.0}}&\textbf{88.0}&88.0&\textbf{90.0}&68.9&79.2&\textbf{75.9}&80.2&76.1\\
		sPGD&89.0&93.0&87.0&\textbf{89.0}&89.5&69.2&\textbf{79.9}&72.8&\textbf{80.6}&75.6\\
		UPGD&79.0&75.0&86.0&69.0&77.3&\textbf{\underline{85.7}}&78.5&74.3&75.2&\textbf{78.4}\\
		\hline
	\end{tabular}}
\end{table*}

In Table~\ref{fr}, we list the fooling rate results on the considered six texture datasets. It can be observed that UAP exhibits a large performance gap in terms of fooling rate as compared to the other attack methods, with a significantly lower performance on the GTOS and DTD datasets. For the two GAP variants, GAP-llc results in significantly lower fooling rates as compared to GAP-tar, while GAP-tar results in the best performances on most targeted models for the MINC, GTOS, DTD and FMD datasets. As for sPGD and UPGD, they produce competitive results that are close on average to the top performing GAP-tar attack method on the considered datasets.

\begin{figure*}[!tb]
	\centering
	\begin{minipage}{0.08\textwidth}
		\centering
		\scriptsize
		\ 
	\end{minipage}
	\begin{minipage}{0.16\textwidth}
		\centering
		\scriptsize
		clean image
	\end{minipage}
	\begin{minipage}{0.16\textwidth}
		\centering
		\scriptsize
		ResNet
	\end{minipage}
	\begin{minipage}{0.16\textwidth}
		\centering
		\scriptsize
		DeepTEN
	\end{minipage}
	\begin{minipage}{0.16\textwidth}
		\centering
		\scriptsize
		DEP
	\end{minipage}
	\begin{minipage}{0.16\textwidth}
		\centering
		\scriptsize
		MuLTER
	\end{minipage}
	\begin{minipage}{0.08\textwidth}
		\centering
		\scriptsize
		\ \\ \ \\
		MINC
		\ \\ \ \\ \ \\ \ \\ \ \\ \ \\ \ \\
		GTOS
		\ \\ \ \\ \ \\ \ \\ \ \\ \ \\ \ \\
		DTD
		\ \\ \ \\ \ \\ \ \\ \ \\ \ \\ \ \\
		4DLF
		\ \\ \ \\ \ \\ \ \\ \ \\ \ \\ \ \\
		FMD
		\ \\ \ \\ \ \\ \ \\ \ \\ \ \\ \ \\
		KTH
	\end{minipage}
	\begin{minipage}{0.16\textwidth}
		\centering
		\includegraphics[width=1\textwidth]{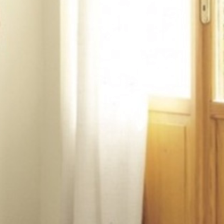}
		\includegraphics[width=1\textwidth]{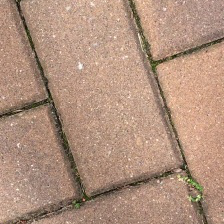}
		\includegraphics[width=1\textwidth]{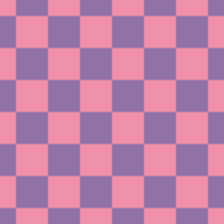}
		\includegraphics[width=1\textwidth]{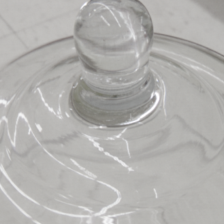}
		\includegraphics[width=1\textwidth]{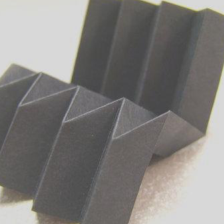}
		\includegraphics[width=1\textwidth]{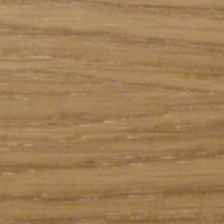}
	\end{minipage}
	\begin{minipage}{0.16\textwidth}
		\centering
		\includegraphics[width=1\textwidth]{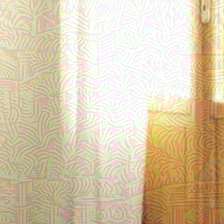}
		\includegraphics[width=1\textwidth]{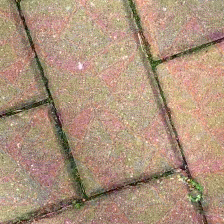}
		\includegraphics[width=1\textwidth]{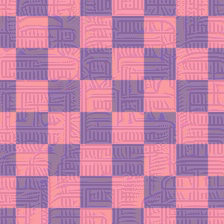}
		\includegraphics[width=1\textwidth]{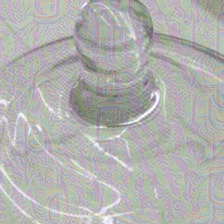}
		\includegraphics[width=1\textwidth]{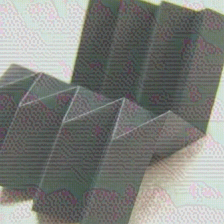}
		\includegraphics[width=1\textwidth]{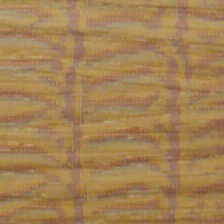}
	\end{minipage}
	\begin{minipage}{0.16\textwidth}
		\centering
		\includegraphics[width=1\textwidth]{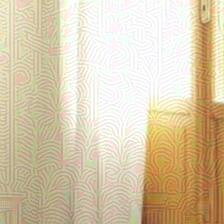}
		\includegraphics[width=1\textwidth]{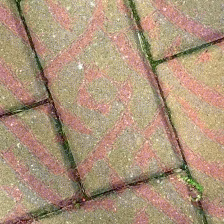}
		\includegraphics[width=1\textwidth]{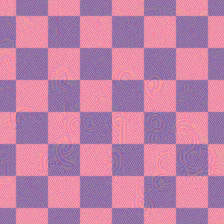}
		\includegraphics[width=1\textwidth]{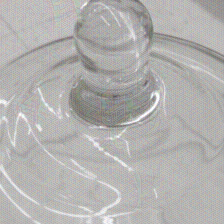}
		\includegraphics[width=1\textwidth]{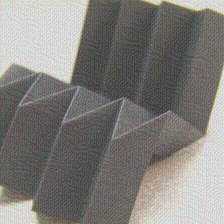}
		\includegraphics[width=1\textwidth]{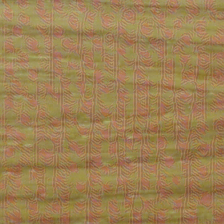}
	\end{minipage}
	\begin{minipage}{0.16\textwidth}
		\centering
		\includegraphics[width=1\textwidth]{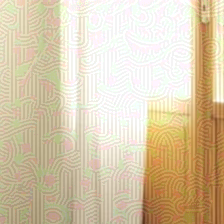}
		\includegraphics[width=1\textwidth]{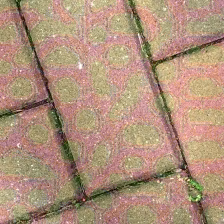}
		\includegraphics[width=1\textwidth]{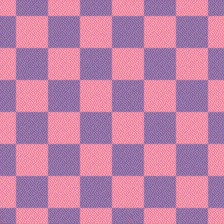}
		\includegraphics[width=1\textwidth]{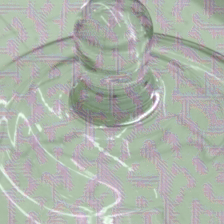}
		\includegraphics[width=1\textwidth]{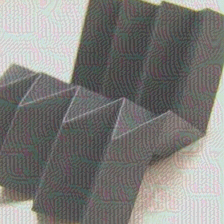}
		\includegraphics[width=1\textwidth]{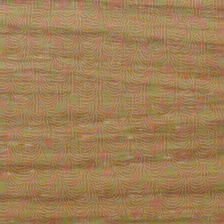}
	\end{minipage}
	\begin{minipage}{0.16\textwidth}
		\centering
		\includegraphics[width=1\textwidth]{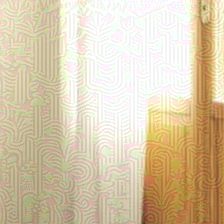}
		\includegraphics[width=1\textwidth]{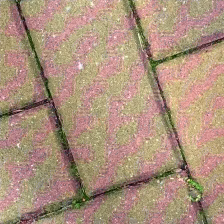}
		\includegraphics[width=1\textwidth]{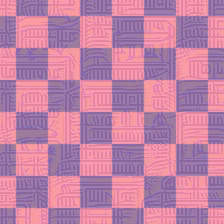}
		\includegraphics[width=1\textwidth]{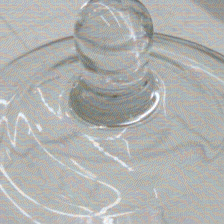}
		\includegraphics[width=1\textwidth]{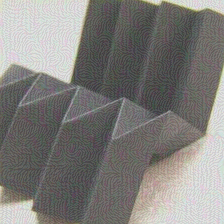}
		\includegraphics[width=1\textwidth]{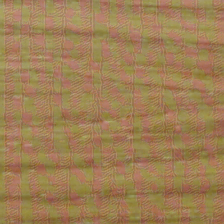}
	\end{minipage}
	\caption{Adversarial examples on different datasets. For each dataset listed in the first column, we show samples of adversarial examples perturbed by GAP-tar against all four targeted models (third to six columns). We also show the original image in the second column as reference.}
	\label{fig:adveg}
\end{figure*}

In addition, GAP-tar, sPGD and UPGD can easily fool all the four models that are trained for the MINC dataset on over 93\% of the testing images. Concerning the vulnerability of the targeted models, attacking DeepTEN yields the highest fooling rates on four (MINC, GTOS, DTD and FMD) out of the six datasets. Overall, the highest fooling rate on each dataset can exceed 80\%, meaning that an effective universal perturbation generally exists for the evaluated texture datasets, even with a significantly small attacking set (including less than 500 images) or few classes.

\begin{figure*}[!tb]
	\centering
	\begin{minipage}{0.08\textwidth}
		\centering
		\scriptsize
		\ 
	\end{minipage}
	\begin{minipage}{0.16\textwidth}
		\centering
		\scriptsize
		clean image
	\end{minipage}
	\begin{minipage}{0.16\textwidth}
		\centering
		\scriptsize
		ResNet
	\end{minipage}
	\begin{minipage}{0.16\textwidth}
		\centering
		\scriptsize
		DeepTEN
	\end{minipage}
	\begin{minipage}{0.16\textwidth}
		\centering
		\scriptsize
		DEP
	\end{minipage}
	\begin{minipage}{0.16\textwidth}
		\centering
		\scriptsize
		MuLTER
	\end{minipage}
	\begin{minipage}{0.08\textwidth}
		\centering
		\scriptsize
		\ \\ \ \\
		MINC
		\ \\ \ \\ \ \\ \ \\ \ \\ \ \\ \ \\
		GTOS
		\ \\ \ \\ \ \\ \ \\ \ \\ \ \\ \ \\
		DTD
		\ \\ \ \\ \ \\ \ \\ \ \\ \ \\ \ \\
		4DLF
		\ \\ \ \\ \ \\ \ \\ \ \\ \ \\ \ \\
		FMD
		\ \\ \ \\ \ \\ \ \\ \ \\ \ \\ \ \\
		KTH
	\end{minipage}
	\begin{minipage}{0.16\textwidth}
		\centering
		\includegraphics[width=1\textwidth]{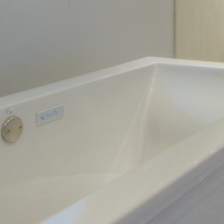}
		\includegraphics[width=1\textwidth]{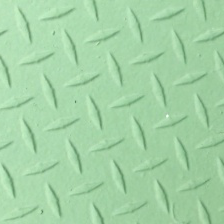}
		\includegraphics[width=1\textwidth]{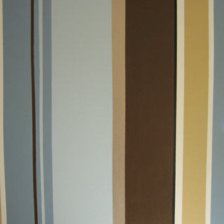}
		\includegraphics[width=1\textwidth]{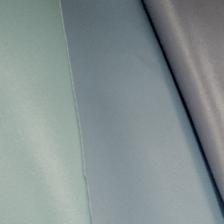}
		\includegraphics[width=1\textwidth]{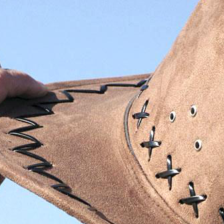}
		\includegraphics[width=1\textwidth]{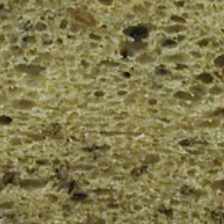}
	\end{minipage}
	\begin{minipage}{0.16\textwidth}
		\centering
		\includegraphics[width=1\textwidth]{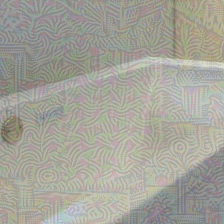}
		\includegraphics[width=1\textwidth]{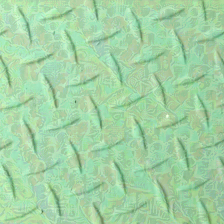}
		\includegraphics[width=1\textwidth]{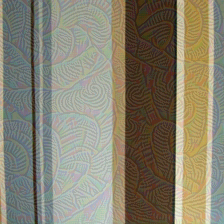}
		\includegraphics[width=1\textwidth]{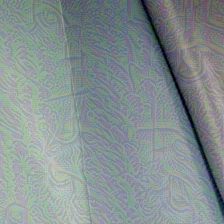}
		\includegraphics[width=1\textwidth]{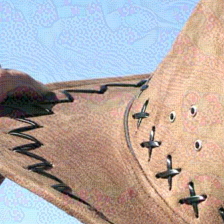}
		\includegraphics[width=1\textwidth]{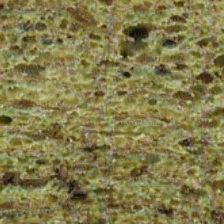}
	\end{minipage}
	\begin{minipage}{0.16\textwidth}
		\centering
		\includegraphics[width=1\textwidth]{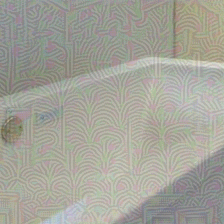}
		\includegraphics[width=1\textwidth]{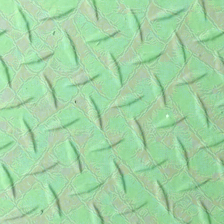}
		\includegraphics[width=1\textwidth]{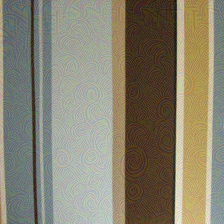}
		\includegraphics[width=1\textwidth]{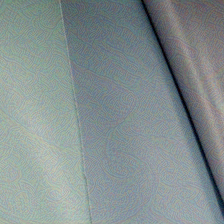}
		\includegraphics[width=1\textwidth]{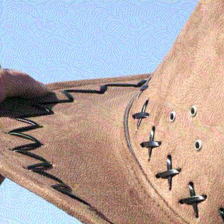}
		\includegraphics[width=1\textwidth]{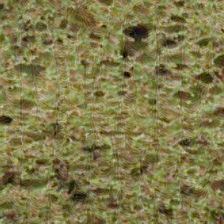}
	\end{minipage}
	\begin{minipage}{0.16\textwidth}
		\centering
		\includegraphics[width=1\textwidth]{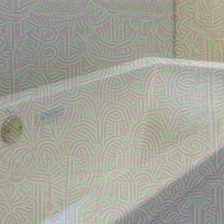}
		\includegraphics[width=1\textwidth]{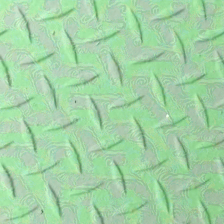}
		\includegraphics[width=1\textwidth]{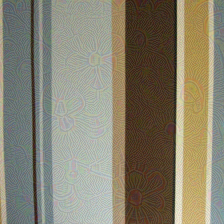}
		\includegraphics[width=1\textwidth]{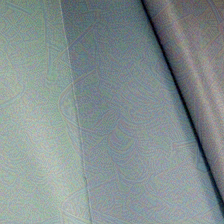}
		\includegraphics[width=1\textwidth]{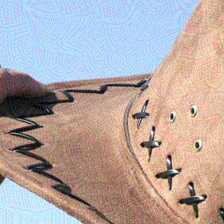}
		\includegraphics[width=1\textwidth]{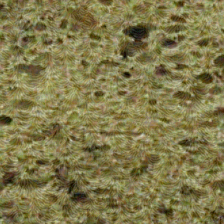}
	\end{minipage}
	\begin{minipage}{0.16\textwidth}
		\centering
		\includegraphics[width=1\textwidth]{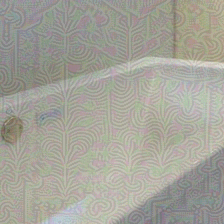}
		\includegraphics[width=1\textwidth]{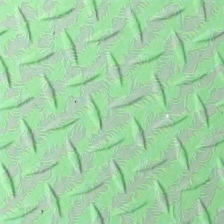}
		\includegraphics[width=1\textwidth]{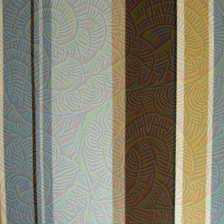}
		\includegraphics[width=1\textwidth]{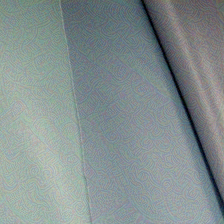}
		\includegraphics[width=1\textwidth]{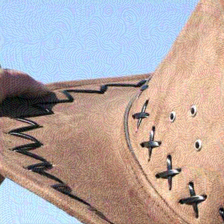}
		\includegraphics[width=1\textwidth]{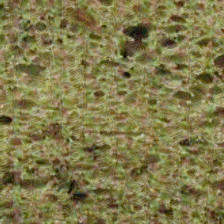}
	\end{minipage}
	\caption{Adversarial examples on different datasets. For each dataset listed in the first column, we show samples of adversarial examples perturbed by sPGD against all four targeted models (third to six columns). We also show the original image in the second column as reference.}
	\label{fig:adveg2}
\end{figure*}

\begin{figure*}[!htb]
	\centering
	\begin{minipage}{0.08\textwidth}
		\centering
		\scriptsize
		\ \\ \ \\
		MINC
		\ \\ \ \\ \ \\ \ \\ \ \\ \ \\ \ \\
		GTOS
		\ \\ \ \\ \ \\ \ \\ \ \\ \ \\ \ \\
		DTD
		\ \\ \ \\ \ \\ \ \\ \ \\ \ \\ \ \\
		4DLF
		\ \\ \ \\ \ \\ \ \\ \ \\ \ \\ \ \\
		FMD
		\ \\ \ \\ \ \\ \ \\ \ \\ \ \\ \ \\
		KTH
	\end{minipage}
	\begin{minipage}{0.16\textwidth}
		\centering
		\scriptsize
		UAP
		\includegraphics[width=1\textwidth]{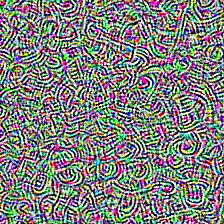}
		\includegraphics[width=1\textwidth]{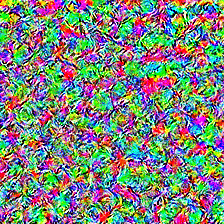}
		\includegraphics[width=1\textwidth]{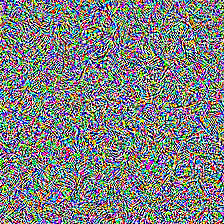}
		\includegraphics[width=1\textwidth]{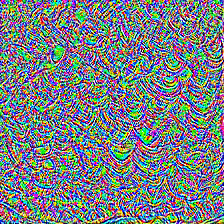}
		\includegraphics[width=1\textwidth]{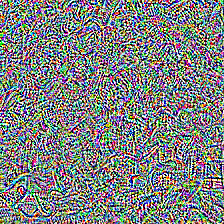}
		\includegraphics[width=1\textwidth]{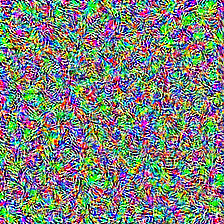}
	\end{minipage}
	\begin{minipage}{0.16\textwidth}
		\centering
		\scriptsize
		GAP-llc
		\includegraphics[width=1\textwidth]{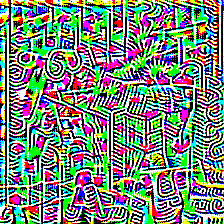}
		\includegraphics[width=1\textwidth]{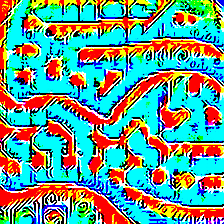}
		\includegraphics[width=1\textwidth]{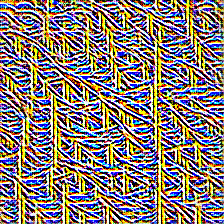}
		\includegraphics[width=1\textwidth]{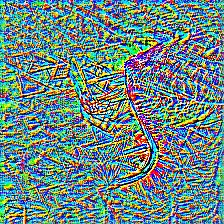}
		\includegraphics[width=1\textwidth]{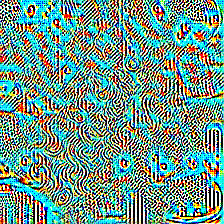}
		\includegraphics[width=1\textwidth]{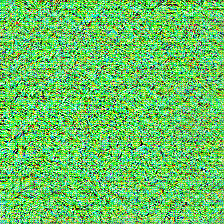}
	\end{minipage}
	\begin{minipage}{0.16\textwidth}
		\centering
		\scriptsize
		GAP-tar
		\includegraphics[width=1\textwidth]{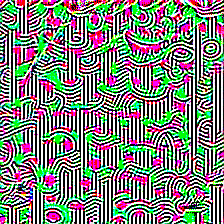}
		\includegraphics[width=1\textwidth]{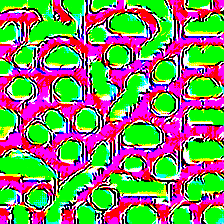}
		\includegraphics[width=1\textwidth]{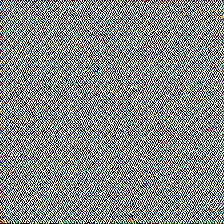}
		\includegraphics[width=1\textwidth]{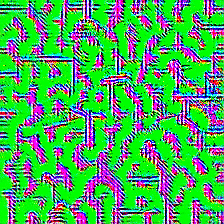}
		\includegraphics[width=1\textwidth]{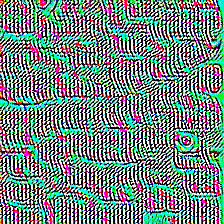}
		\includegraphics[width=1\textwidth]{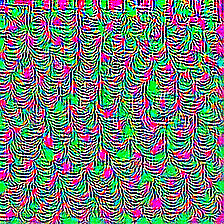}
	\end{minipage}
	\begin{minipage}{0.16\textwidth}
		\centering
		\scriptsize
		sPGD
		\includegraphics[width=1\textwidth]{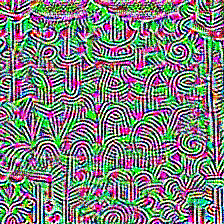}
		\includegraphics[width=1\textwidth]{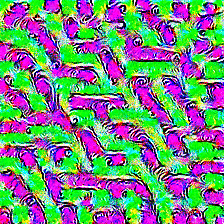}
		\includegraphics[width=1\textwidth]{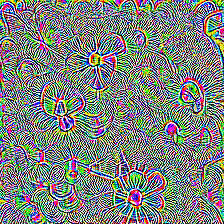}
		\includegraphics[width=1\textwidth]{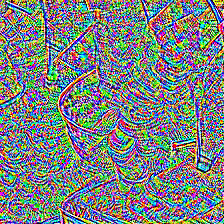}
		\includegraphics[width=1\textwidth]{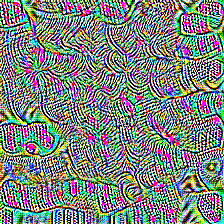}
		\includegraphics[width=1\textwidth]{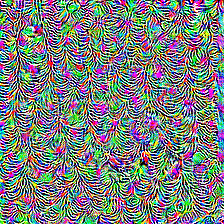}
	\end{minipage}
	\begin{minipage}{0.16\textwidth}
		\centering
		\scriptsize
		UPGD
		\includegraphics[width=1\textwidth]{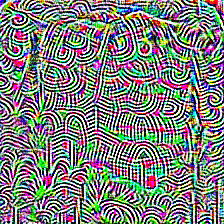}
		\includegraphics[width=1\textwidth]{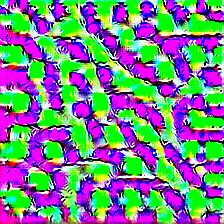}
		\includegraphics[width=1\textwidth]{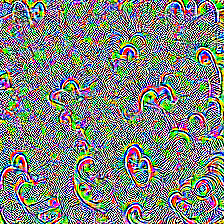}
		\includegraphics[width=1\textwidth]{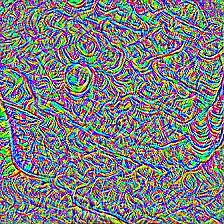}
		\includegraphics[width=1\textwidth]{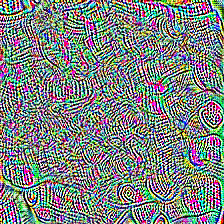}
		\includegraphics[width=1\textwidth]{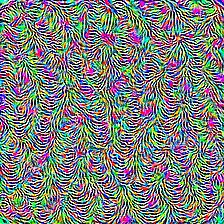}
	\end{minipage}
	\caption{Visualization of perturbations computed using different universal attack methods (first row) against DEP models on the six datasets (first column). The perturbations are rescaled for visibility.}
	\label{fig:perteg}
\end{figure*}

\subsubsection{Visual Examples}

Figure~\ref{fig:adveg} and~\ref{fig:adveg2} show, respectively, some visual examples of images perturbed by GAP-tar and sPGD against four models for each dataset (Section~\ref{attack}). Typically, perturbations are quasi-imperceptible given the relatively small $l_\infty$ norm. The perturbations for the GTOS dataset consist of colored patterns and blobs, which are more perceivable than the other types of patterns. According to Figure~\ref{fig:perteg}, most of the perturbations show dense texture patterns, which are considered as middle/high frequency components, while some perturbations especially for the GTOS dataset also include colored structured patterns (e.g., spots and strips).

\section{Conclusion}

To evaluate the robustness of deep learning models for texture recognition against universal attacks, we train four deep learning models (ResNet, DeepTEN, DEP and MuLTER) on six texture datasets (MINC, GTOS, DTD, 4DLF, FMD and KTH) separately and conduct five existing universal attack methods (UAP, GAP-llc, GAP-tar, sPGD and UPGD) by targeting all the models. The observed results show that highly effective universal perturbations exist over all tested datasets regardless of training data size and class number, achieving fooling rates of more than 80\%. We also observe that these perturbations contain various components like colored flat regions (low frequency), intense lines (middle frequency) and dense textures (high frequency).

\bibliographystyle{IEEEbib}
\bibliography{refs}

\end{document}